\documentclass{article}
\usepackage[
    backend=biber,
    style=numeric,
    sorting=none
]{biblatex}
\usepackage{hyperref}
\usepackage[english]{babel}
\usepackage[letterpaper,top=2cm,bottom=2cm,left=3cm,right=3cm,marginparwidth=1.75cm]{geometry}
\usepackage[table,xcdraw]{xcolor}
\usepackage{amsmath}
\usepackage{braket}
\usepackage{titlesec}
\usepackage[usestackEOL]{stackengine}
\usepackage{array}
\usepackage{wrapfig}
\usepackage{tikz}
\usepackage{caption}
\usepackage{graphicx}
\usepackage{subcaption}
\usepackage{bm}	
\usepackage{color}                     
\usepackage{xcolor}
\usepackage{epsfig}
\usepackage{amsmath} 
\usepackage{amssymb} 
\usepackage{float}
\usepackage{mathtools}
\usepackage{xparse}
\usepackage{times}
\usepackage{subcaption}
\usepackage[normalem]{ulem}

\usetikzlibrary{quantikz}
\graphicspath{ {./images/} }

\renewcommand{\eqref}[1]{Eq. ({\ref{#1}})}

\title{\Large From hyperplanes to hyperellipsoids: a pedagogical bridge from linear classifiers to single-qubit mixed-state classifiers}

\author{\large Kaitlin Gili$^{1}$ \\[0.2em] {\small $^{1}$QodeX Quantum}} \date{\small July 9, 2026}
\begin{document}
\maketitle

\vspace{-2em}

\begin{abstract} 
\normalsize

Quantum machine learning ideas are typically absent from traditional machine learning curricula and are instead introduced in upper-level courses on quantum information processing and quantum computing. We propose a route for bringing quantum ideas into the standard machine learning classroom directly after the introduction of binary linear classifiers. Specifically, we develop a pedagogical bridge between linear classifiers and single-qubit mixed-state classifiers by characterizing their geometric model structure. This characterization reveals that a single-qubit mixed-state model for binary classification can be viewed as the ``ellipsoidal version'' of a standard linear classifier: rather than learning a \emph{hyperplane} that separates data, the model learns a \emph{hyperellipsoid}. We present this connection as an accessible bridge from familiar classical machine learning concepts to quantum machine learning, and encourage instructors to use it as a natural entry point for introducing quantum ideas in the undergraduate machine learning classroom.
\end{abstract}

\section{Introduction}

We're interested in supporting machine learning (ML) pedagogy by providing instructors with natural entry points for introducing quantum ML ideas into the classroom. In this brief note, we establish a pedagogical connection that allows a single qubit mixed state model \cite{nielsen2010quantum} to be smoothly introduced following the standard linear classifier \cite{Bishop-book-2006, hastie2009elements, james2021introduction}. Linear classification is arguably the most widely used ML method across industry sectors -- largely for its simplicity and inherent interpretability \cite{fan2008liblinear}. Often-times these models are implemented as last layer classification heads for deep neural networks \cite{he2016deep, BERT}, or as classifiers for frozen embeddings from a foundation model, known as linear probing \cite{alain2016understanding, evci2022head2toe}. Occasionally, these models even perform on par with deep non-linear ones \cite{schulz2020different}, or outperform them entirely for a specific problem \cite{arshad2023powersimplicitysimplelinear}. Linear classification is typically among the first parametric methods taught in an undergraduate introductory ML course \cite{james2021introduction}. On the opposite side of the spectrum, a single quantum bit (qubit) mixed-state model is typically viewed as requiring specialized knowledge of quantum information science to understand and use \cite{nielsen2010quantum}. The model name contains the word ``qubit" -- which for many people immediately triggers a notion of inaccessibility. In fact, students may only learn about this model in a graduate quantum information course -- completely separate from ML methods. 

We show that if we look past the complex jargon, a single qubit mixed state model for binary classification is just the ``ellipsoid version" of standard linear model classification. More precisely, rather than learning a \emph{hyperplane} to classify data, we learn a \emph{hyperellipsoid}. We characterize this difference in geometric inductive bias between the models, as well as the effect of the additional restriction imposed by the normalization constraint on the weights in the qubit case. In the latter bias, we observe that linear models offer inherent interpretability through \emph{absolute} feature importance, whereas single-qubit mixed-state models offer \emph{relative} feature importance.

We hope this work encourages instructors to bring quantum ML ideas into standard ML courses in a way that demystifies quantum models by showing how they differ from familiar classical models through their mathematical structure.

\section{Introducing a pedagogical bridge between classifiers}

\begin{wrapfigure}{r}{0.65\textwidth}
    \centering
    \vspace{-0.5em}
    \begin{subfigure}[b]{0.5\linewidth}
        \centering
        \includegraphics[width=\linewidth]{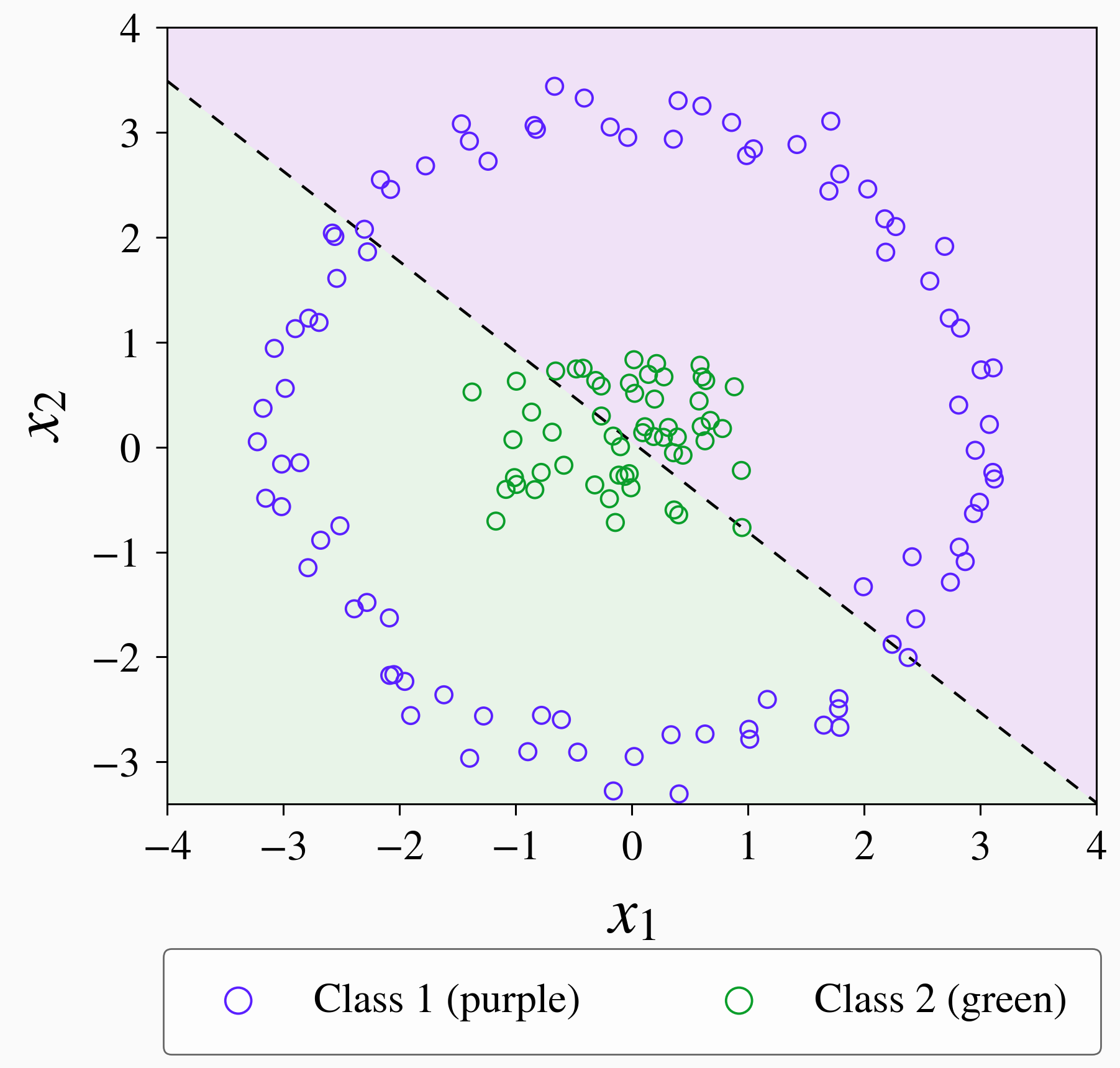}
        \caption{Linear model}\label{figure1}
    \end{subfigure}
    \hspace{-0.5em}
    \begin{subfigure}[b]{0.5\linewidth}
        \centering
        \includegraphics[width=\linewidth]{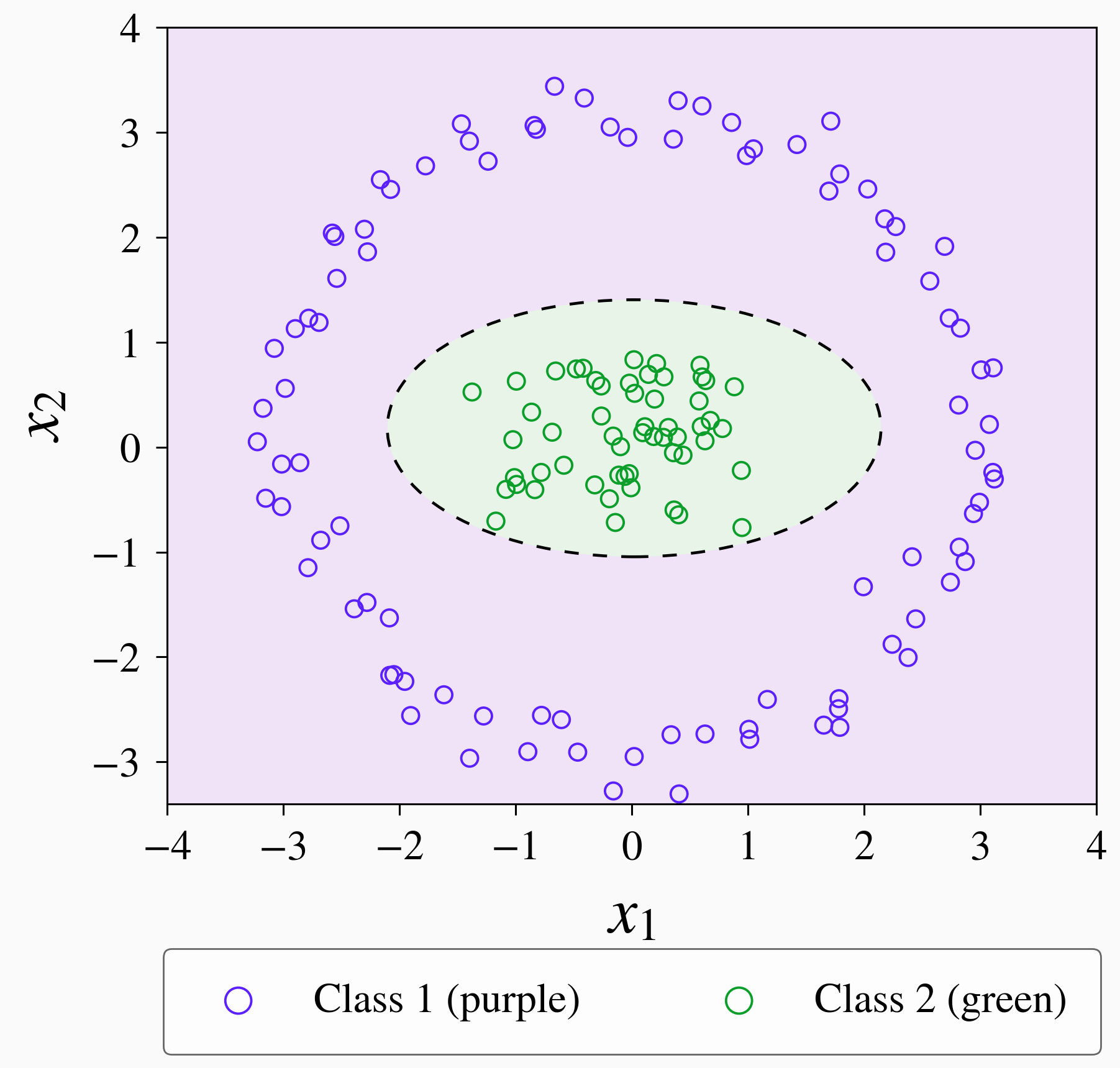}
        \caption{Single qubit mixed state model}\label{figure2}
    \end{subfigure}
    \caption{Toy classification example for two data features. (a) The model can only learn the slope of a line to separate the data. (b) The model can only learn the semi-axis lengths of an ellipsoid to separate the data.}
    \label{model-comparison}
\end{wrapfigure}

Let us consider a traditional supervised learning classification problem. We have a training dataset $\mathcal{D} = \{x_i, y_i\}_{i=1}^N$, where $x_i \in \mathbb{R}^D$ and $y_i \in \{0,1\}$. Our goal is to train a parameterized model to classify the input data $x_i$ by its true binary label $y_i = 0$ or $y_i = 1$, and generalize to unseen data that is assumed to be sampled from the same underlying distribution as the training set. Typically, we train such a model with a logistic loss function. Below, we characterize the interpretability offerings of linear models and single qubit mixed state models for this task. 

\subsection{Linear model}
Let us consider a linear classification model. For a single datapoint $x_i$, a linear model is of the form: 

\begin{equation}
    f_{w}({x}) =  w_1x_{1} + w_2x_{2} + \dots + w_Dx_{D} + \beta \label{linear_model}
\end{equation}

Each trainable weight $w_d \in \mathbb{R}$ influences a data feature $x_d$ directly, and the combined sum is the model output for the entire data input $x_i$, plus a scalar bias offset $\beta \in \mathbb{R}$. In this form, the parameterized weights are unbounded in training -- although constraints can be added. If raw data features are very large compared to others, we use a $\min-\max$ normalization to map each datapoint $x_i \in [0,1]$. This normalization makes the model output more sensitive to the learned feature weights than to the absolute scale of the input data. We set the binary decision threshold to zero, and so the final decision function is $\text{sign}(f_{w}({x}))$. 

\noindent \textbf{Absolute feature importance.} The sign (positive or negative) of each feature weight $w_d$ indicates whether that feature pushes the model output toward class $0$ or class $1$. The magnitude $|w_d|$ indicates \emph{absolute importance} of the feature weight in its contribution to the classification output. Thus, we understand what features have high or low impact on the model behavior. 

\noindent \textbf{Learning a hyperplane.} When $f_{w}({x}) = 0$, we have a $D-$dimensional hyper-plane, and thus we are learning this hyper-plane that separates the data. The weights determine the orientation and placement of the hyper-plane. Linear models have the constraint that they can \emph{only} learn linear decision boundaries. Hence, if the data cannot be linearly separated, a linear model is not geometrically expressive enough for perfect classification, no matter how much additional data is available. See Fig.~\ref{figure1} for a standard toy example with two data features. Conversely, if we know that the data are linearly separable, why use anything more complicated?

\subsection{Single qubit mixed state model}

Let us now consider a single qubit mixed state model \footnote{As this is a \emph{single} qubit model, it is clearly classically simulatable. It is useful to classify this model as \emph{quantum-inspired} for the reason that we would not actually need to use quantum hardware resources to implement it.}. In order to reveal the hyperellipsoid structure of the model, we need to introduce two ideas within quantum information science -- a qubit state and a qubit measurement \cite{nielsen2010quantum}. However, we encourage the reader to not pay close attention to the physical meaning of these ideas (e.g.~the physical state of an atom and our observation of it via interaction); rather, it is important that one pay attention to the mathematical objects that represent the physical ones (i.e.~a unit vector and a Hermitian matrix). 

A \emph{pure} single qubit state is a $2-$dimensional column unit vector containing complex coefficients $\alpha \in \mathbb{C}$ and $\beta \in \mathbb{C}$. We denote this as $v_{\text{pure}} = \begin{pmatrix}
\alpha \\
\beta
\end{pmatrix}$. Given that $v_{\text{pure}}$ is a unit vector, we know that the coefficients are constrained as $|\alpha|^2 + |\beta|^2 =1$. The coefficients indicate rotational coordinates of the vector on the unit sphere. 

A \emph{mixed} single qubit state is a linear sum of pure model outer products, where each is multiplied by some probability $p$. We encode each data feature $d$ in a datapoint $x_i$ into the $\alpha$ coefficient of each pure state. This technique is a form of amplitude encoding \cite{grover2002creating, Rebentrost-PRL-2014}. Since all of our data are real-valued, from this point onward we restrict $\alpha$ to be real. Each probability $p_d$ is the trainable feature weight, analogous to the weight $w$ above. As $\alpha^2 < 1$, we use a $\min-\max$ normalization to map all data features $x_d \in [0,1]$ (same as in the linear model). Under these conditions, we observe the following single qubit mixed state: 

\begin{equation}
    v_{\text{mixed}} = \sum _{d=1}^D p_d \ v_{\text{pure}}(x_d)v_{\text{pure}}^T(x_d) \\
= \sum_{d=1}^D p_d\begin{pmatrix}
|x_d|^2 & x_d\beta\\
\beta x_d & |\beta|^2 \\ \end{pmatrix} \\
= \sum_{d=1}^D p_d\begin{pmatrix}
|x_d|^2 & x_d\beta\\
\beta x_d & 1 - |x_d|^2 \\ \end{pmatrix}
\end{equation}

Notice that the diagonals can be written in this form because $|\alpha|^2 + |\beta|^2 =1$. As such, quantum information imposes the general normalization constraint $\text{Tr}(v_{\text{mixed}})=1$.

So far, we have the mixed quantum state $v_{\text{mixed}} \in \mathbb{R}^{2 \times 2}$, but not the full single qubit mixed state model. The full model includes a measurement of this quantum state with respect to a chosen random variable (e.g.~system energies). The expectation of measuring a particular random variable is the trace (sum of the diagonal entries) of the product of the state $v_{\text{mixed}}$ and the possible random variable outcomes represented as the Hermitian (real-eigenvalued) matrix $\mathcal{O} \in \mathbb{C}^{2 \times 2} $. 

If we choose a standard binary computational measurement $Z = \begin{pmatrix}
1 & 0\\
0 & -1 \end{pmatrix}$, the single qubit mixed state model is written as:  
\begin{equation}
\begin{aligned}
f_p(x)
&= \operatorname{Tr}\!\left[v_{\text{mixed}} \mathcal{O}\right] \\
&= \operatorname{Tr}\!\left[
\sum_{d=1}^D p_d
\begin{pmatrix}
|x_d|^2 & x_d\beta \\
\beta x_d & 1-|x_d|^2
\end{pmatrix}
\begin{pmatrix}
1 & 0 \\
0 & -1
\end{pmatrix}
\right] \\
&= \operatorname{Tr}\!\left[
\sum_{d=1}^D p_d
\begin{pmatrix}
|x_d|^2 & -x_d\beta \\
\beta x_d & |x_d|^2-1
\end{pmatrix}
\right] \\
&= 2\left(
p_1|x_1|^2
+ p_2|x_2|^2
+ \dots
+ p_D|x_D|^2
\right)-1
\end{aligned}\label{single_qubit_mixed_model}
\end{equation}

Each trainable weight $p_d$ directly scales the squared contribution of the corresponding input feature $x_d$, and the combined sum is the model output for the entire data input $x_i$. In this standard form, the parameterized weights \emph{are bounded} in training -- unlike the standard linear model. The probabilities are bounded as $0 \leq p_d \leq 1$. We set the binary decision threshold to zero, and so the final decision function is $\text{sign}(f_{p}({x}))$. 

\noindent \textbf{Relative feature importance.} All probabilistic weights are positive, and thus the sign of each weight provides no information regarding its contribution to the classification outcome. However, the constraint $\sum_{d=1}^D p_d =1$ adds an inherently interpretable bias of relative importance for each feature's contribution to the model output. Thus, the model encodes the bias that some features are more important than others -- analogous to the bias of self-attention in transformer architectures. 

\noindent \textbf{Learning a hyperellipsoid.} Observe that the single qubit mixed state model takes a final form that is similar to the linear model in Eq.~\ref{linear_model}, except that rather than containing the structure of linear hyper-plane, the model structure is that of a \emph{hyperellipsoid}. When the decision threshold is zero, we obtain the following hyperellipsoid boundary, centered at the origin: 
\begin{equation}
    \frac{x_1^2}{a_1^2}
    + \frac{x_2^2}{a_2^2}
    + \dots
    + \frac{x_D^2}{a_D^2}
    = 1
\end{equation}
The coefficients $a_d$ are semi-axis lengths -- i.e.~the distances from the origin to the ellipsoid's boundary along each direction. From our model in Eq.~\ref{single_qubit_mixed_model}, we observe that $a_d = \frac{1}{\sqrt{2p_d}}$. As a result, changing each probability is geometrically equivalent to stretching or shrinking the ellipsoid along a distinct feature direction. See Fig.~\ref{model-comparison} for a comparison between both models with toy data when $D=2$. When all of the data features are of equal relative importance ($p_d=\frac{1}{D}$), we obtain a \emph{hyper-sphere} around the origin with a radius $r = \sqrt{\frac{D}{2}}$. The hyperellipsoid is most compressed along a feature direction when $p_d=1$ and stretches without bound as $p_d\to 0$. When $p_d=0$, the corresponding feature no longer contributes to the decision boundary, causing the hyperellipsoid to become unbounded in that direction and geometrically reducing it to a hypercylinder. Thus, from a practical implementation standpoint, it is important to lower-bound the probabilistic weights by a small positive value, such as $10^{-9}$, to prevent the decision boundary from becoming unbounded along any feature direction. Similar to the linear model, the single qubit mixed state model can only learn decision boundaries that are ellipsoidal. Fig.~\ref{figure1} presents a toy setting in which this ellipsoidal boundary can separate data that a linear classifier cannot. While this provides greater curvature around the data, the model remains more restricted than one that could also translate the hyperellipsoid away from the origin.

For the reader who is interesting in learning more about single qubit mixed state models for ML, we refer them to Refs.~\cite{sergioli2016pattern, leporini2022efficient, bertini2023quantum}. In these different Refs, the authors explore models where the mixed state probabilities are fixed, and individual datapoints are encoded into each pure state to be classified via distance metrics. Alternatively, for readers interested in the more widely studied single-qubit quantum classifier, we refer them to Ref.~\cite{perez2020reuploading}. This method classifies data by encoding each input into a single pure quantum state, training the encoding via unitary rotations, and then computing the expectation with respect to the random variable $Z$. Lastly, if one is interested in different ML problems that can be tackled with only a single qubit, we refer the reader to Ref.~\cite{cuellar2024onequbit}. To the best of our knowledge, the exact single qubit mixed model discussed here, particularly the feature of the probabilities themselves being trainable, has not been studied in other works. At least, this model does not seem to be present in the mainstream quantum ML literature. While we introduce the model characterization to support ML pedagogy,  we encourage interested researchers and practitioners to investigate the classifier's empirical behavior on real-world data. 

\section{Conclusion}
By characterizing models through their inherent structure, and discussing their differences in a form of analysis that is closer to \emph{qualitative} than quantitative, we are able to draw pedagogical connections between models that may once have seemed far apart. 
For machine learning instructors, specifically, we hope our characterization above provides a way to bring quantum ideas into the classroom that is only one step removed from linear classification. Furthermore, for quantum information instructors, perhaps it offers a natural route for introducing machine learning ideas alongside qubits and mixed states, while providing an intuitive geometric perspective. And lastly for anyone who has viewed quantum models as somewhat ``fuzzy'' objects, with little clarity about how they relate to familiar classical models, we hope this perspective makes quantum ideas feel closer and more concrete.

\printbibliography

@inproceedings{evci2022head2toe,
  author    = {Evci, Utku and Dumoulin, Vincent and Larochelle, Hugo and Mozer, Michael C.},
  title     = {{Head2Toe}: Utilizing Intermediate Representations for Better Transfer Learning},
  booktitle = {Proceedings of the 39th International Conference on Machine Learning},
  editor    = {Chaudhuri, Kamalika and Jegelka, Stefanie and Song, Le and Szepesv{\'a}ri, Csaba and Niu, Gang and Sabato, Sivan},
  series    = {Proceedings of Machine Learning Research},
  volume    = {162},
  pages     = {6009--6033},
  year      = {2022},
  publisher = {PMLR},
  url       = {https://proceedings.mlr.press/v162/evci22a.html}
}

@article{bertini2023quantum,
  title={Quantum-Inspired Applications for Classification Problems},
  author={Bertini, Cesarino and Leporini, Roberto},
  journal={Entropy},
  volume={25},
  number={3},
  pages={404},
  year={2023},
  publisher={MDPI},
  doi={10.3390/e25030404}
}

@article{leporini2022efficient,
  title={An efficient geometric approach to quantum-inspired classifications},
  author={Leporini, Roberto and Pastorello, Davide},
  journal={Scientific Reports},
  volume={12},
  number={1},
  pages={8781},
  year={2022},
  publisher={Nature Publishing Group},
  doi={10.1038/s41598-022-12392-1}
}

@inproceedings{he2016deep,
  author    = {He, Kaiming and Zhang, Xiangyu and Ren, Shaoqing and Sun, Jian},
  title     = {Deep Residual Learning for Image Recognition},
  booktitle = {Proceedings of the IEEE Conference on Computer Vision and Pattern Recognition},
  pages     = {770--778},
  year      = {2016},
  url       = {https://doi.org/10.1109/CVPR.2016.90}
}

@article{alain2016understanding,
  author        = {Alain, Guillaume and Bengio, Yoshua},
  title         = {Understanding Intermediate Layers Using Linear Classifier Probes},
  journal       = {arXiv preprint arXiv:1610.01644},
  year          = {2016},
  eprint        = {1610.01644},
  archiveprefix = {arXiv},
  primaryclass  = {stat.ML},
}

@misc{arshad2023powersimplicitysimplelinear,
      title={The Power Of Simplicity: Why Simple Linear Models Outperform Complex Machine Learning Techniques -- Case Of Breast Cancer Diagnosis}, 
      author={Muhammad Arbab Arshad and Sakib Shahriar and Khizar Anjum},
      year={2023},
      eprint={2306.02449},
      archivePrefix={arXiv},
      primaryClass={cs.LG},
}

@article{schulz2020different,
  author    = {Schulz, Marc-Andr{\'e} and Yeo, B. T. Thomas and Vogelstein, Joshua T. and Mourao-Miranda, Janaina and Kather, Jakob Nikolas and Kording, Konrad and Richards, Blake and Bzdok, Danilo},
  title     = {Different Scaling of Linear Models and Deep Learning in {UK Biobank} Brain Images versus Machine-Learning Datasets},
  journal   = {Nature Communications},
  volume    = {11},
  number    = {1},
  pages     = {4238},
  year      = {2020},
  month     = aug,
  publisher = {Nature Publishing Group},
  url       = {https://doi.org/10.1038/s41467-020-18037-z}
}

@misc{BERT,
      title={BERT: Pre-training of Deep Bidirectional Transformers for Language Understanding}, 
      author={Jacob Devlin and Ming-Wei Chang and Kenton Lee and Kristina Toutanova},
      year={2019},
      archivePrefix={arXiv},
      primaryClass={cs.CL},
      eprint={1810.04805}, 
}

@article{grover2002creating,
  title         = {Creating Superpositions That Correspond to Efficiently Integrable Probability Distributions},
  author        = {Grover, Lov and Rudolph, Terry},
  year          = {2002},
  eprint        = {quant-ph/0208112},
  archivePrefix = {arXiv},
  primaryClass  = {quant-ph}
}

@article{fan2008liblinear,
  author  = {Fan, Rong-En and Chang, Kai-Wei and Hsieh, Cho-Jui and Wang, Xiang-Rui and Lin, Chih-Jen},
  title   = {{LIBLINEAR}: A Library for Large Linear Classification},
  journal = {Journal of Machine Learning Research},
  volume  = {9},
  number  = {61},
  pages   = {1871--1874},
  year    = {2008},
  url     = {https://www.jmlr.org/papers/v9/fan08a.html}
}

@article{cuellar2024onequbit,
  author    = {Cu{\'e}llar, Manuel P.},
  title     = {What We Can Do with One Qubit in Quantum Machine Learning: Ten Classical Machine Learning Problems That Can Be Solved with a Single Qubit},
  journal   = {Quantum Machine Intelligence},
  volume    = {6},
  number    = {2},
  pages     = {76},
  year      = {2024},
  month     = nov,
  publisher = {Springer},
  url       = {https://doi.org/10.1007/s42484-024-00210-y}
}

@article{sergioli2016pattern,
  title   = {A Quantum-Inspired Version of the Nearest Mean Classifier},
  author  = {Sergioli, Giuseppe and Santucci, Enrica and Didaci, Luca and Miszczak, Jaros{\l}aw Adam and Giuntini, Roberto},
  journal = {Soft Computing},
  volume  = {22},
  number  = {3},
  pages   = {691--705},
  year    = {2018},
  publisher = {Springer},
  url     = {https://link.springer.com/article/10.1007/s00500-016-2478-2}
}

@book{hastie2009elements,
  author    = {Hastie, Trevor and Tibshirani, Robert and Friedman, Jerome},
  title     = {The Elements of Statistical Learning: Data Mining, Inference, and Prediction},
  edition   = {2},
  year      = {2009},
  series    = {Springer Series in Statistics},
  publisher = {Springer},
  address   = {New York, NY},
  isbn      = {978-0-387-84857-0},
  url       = {https://doi.org/10.1007/978-0-387-84858-7}
}

@book{james2021introduction,
  title     = {An Introduction to Statistical Learning: with Applications in R},
  author    = {James, Gareth and Witten, Daniela and Hastie, Trevor and Tibshirani, Robert},
  year      = {2021},
  edition   = {2},
  publisher = {Springer},
  address   = {New York},
  series    = {Springer Texts in Statistics},
  doi       = {10.1007/978-1-0716-1418-1}
}

@book{nielsen2010quantum,
  title     = {Quantum Computation and Quantum Information},
  author    = {Nielsen, Michael A. and Chuang, Isaac L.},
  year      = {2010},
  edition   = {10th Anniversary},
  publisher = {Cambridge University Press},
  address   = {Cambridge},
  isbn      = {978-1-107-00217-3},
  doi       = {10.1017/CBO9780511976667}
}

@book{Bishop-book-2006,
  author    = {Bishop, Christopher M.},
  title     = {Pattern Recognition and Machine Learning},
  year      = {2006},
  series    = {Information Science and Statistics},
  publisher = {Springer},
  address   = {New York, NY},
  isbn      = {978-0-387-31073-2},
  url       ={https://link.springer.com/book/9780387310732}
}

@article{Rebentrost-PRL-2014,
  author    = {Rebentrost, Patrick and Mohseni, Masoud and Lloyd, Seth},
  title     = {Quantum Support Vector Machine for Big Data Classification},
  journal   = {Physical Review Letters},
  volume    = {113},
  number    = {13},
  pages     = {130503},
  year      = {2014},
  month     = sep,
  publisher = {American Physical Society},
  url       = {https://doi.org/10.1103/PhysRevLett.113.130503}
}

@article{perez2020reuploading,
  author    = {P{\'e}rez-Salinas, Adri{\'a}n and Cervera-Lierta, Alba and Gil-Fuster, Elies and Latorre, Jos{\'e} I.},
  title     = {Data Re-Uploading for a Universal Quantum Classifier},
  journal   = {Quantum},
  volume    = {4},
  pages     = {226},
  year      = {2020},
  month     = feb,
  publisher = {Verein zur F{\"o}rderung des Open Access Publizierens in den Quantenwissenschaften},
  url       = {https://doi.org/10.22331/q-2020-02-06-226}
}
\end{document}